\pdfoutput=1

\documentclass[11pt]{article}

\usepackage[]{acl}
\usepackage{times}
\usepackage{latexsym}
\usepackage{graphicx}

\usepackage{color}
\usepackage{booktabs}
\usepackage{multicol}
\usepackage{graphicx}
\usepackage{xcolor}
\usepackage{caption}
\usepackage{subcaption}
\usepackage{wrapfig}
\usepackage{colortbl}
\usepackage{bm}
\usepackage{amssymb}
\usepackage{soul}
\usepackage{tabularx}
\usepackage{float}
\usepackage[htt]{hyphenat}
\usepackage{tabu}
\usepackage{amsmath,bm}
\usepackage{array,multirow,colortbl}
\usepackage{todonotes}
\usepackage{multirow}
\usepackage{float}

\usepackage{placeins}

\usepackage{microtype}

\usepackage{times}
\usepackage{latexsym}

\usepackage[T1]{fontenc}

\usepackage[utf8]{inputenc}

\usepackage{microtype}

\title{Partial-input baselines show that NLI models can \\
ignore context, but they don't.}

\author{Neha Srikanth \\
  Department of Computer Science \\
  University of Maryland, College Park \\
  \texttt{nehasrik@umd.edu} \\\And
  Rachel Rudinger \\
  Department of Computer Science \\
  University of Maryland, College Park \\
  \texttt{rudinger@umd.edu} \\}

\begin{document}
\maketitle
\begin{abstract}
When strong partial-input baselines reveal artifacts in crowdsourced NLI datasets, the performance of full-input models trained on such datasets is often dismissed as reliance on spurious correlations. We investigate whether state-of-the-art NLI models are capable of overriding default inferences made by a partial-input baseline. We introduce an evaluation set of 600 examples consisting of perturbed premises to examine a RoBERTa model's sensitivity to edited contexts. Our results indicate that NLI models are still capable of learning to condition on context---a necessary component of inferential reasoning---despite being trained on artifact-ridden datasets.
\end{abstract}

\definecolor{lightteal}{RGB}{208,223,226}
\definecolor{teal}{RGB}{69,129,129}

\definecolor{lightorange}{RGB}{252,229,205}
\definecolor{burntorange}{RGB}{207,146,82}

\definecolor{lightpurple}{RGB}{217,210,233}
\definecolor{darkpurple}{RGB}{124,102,179}

\section{Introduction}
Natural language inference (NLI) is integral to building systems that are broadly capable of language understanding \cite{white2017inference}. In a traditional NLI setup, models are provided with a \textit{premise} as context and a corresponding \textit{hypothesis}. They must then determine whether the premise entails, contradicts, or is neutral in relation to the hypothesis \cite{giampiccolo2007third}. 

Researchers have shown that many NLI datasets contain statistical biases, or ``annotation artifacts'' \cite{gururangan2018annotation, herlihy2021mednli} that systems leverage to correctly predict entailment. To diagnose such artifacts in datasets and provide a stronger alternative to majority-class baselines, \citet{poliak2018hypothesis} introduced partial-input baselines, a setting in which models are provided parts of NLI instances to predict an entailment relation. \citet{poliak2018hypothesis}, \citet{gururangan2018annotation}, and others posit that datasets containing artifacts may in turn produce models that are incapable of learning to perform true reasoning.

\begin{figure}[t]
\centering
\includegraphics[scale=0.60]{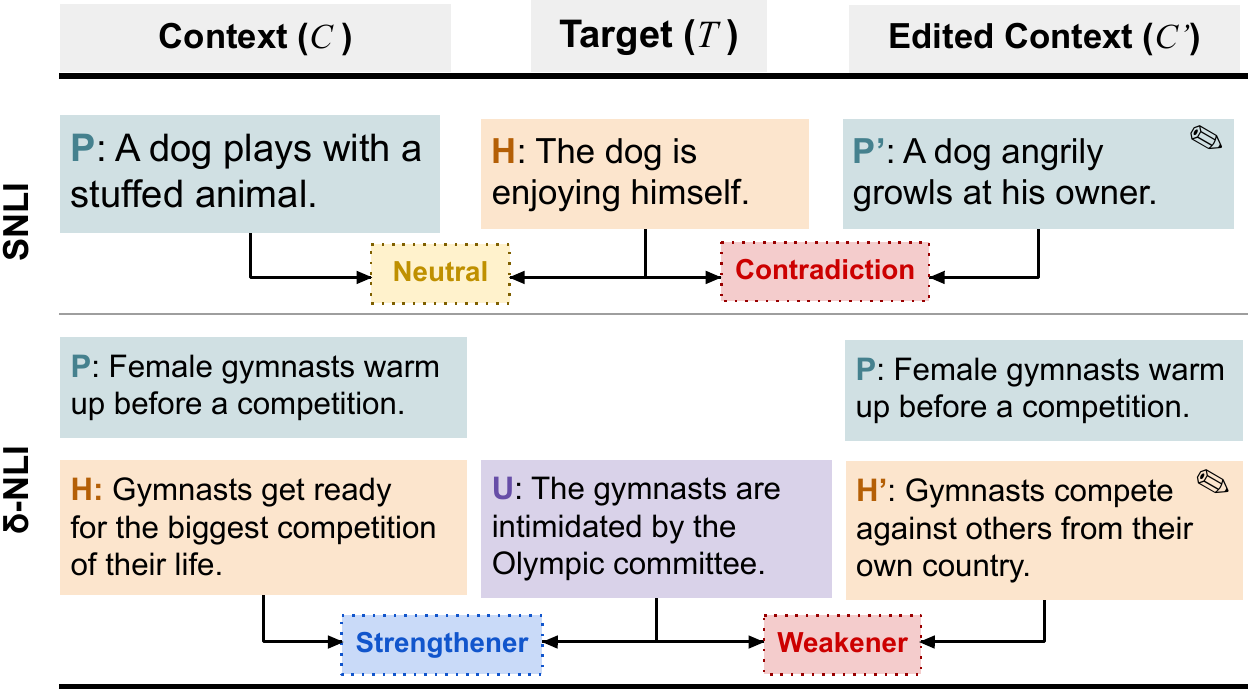}
\caption{Given NLI and $\delta$-NLI instances that partial-input baselines correctly label, we edit their context to induce a different gold label. We use the edited instances to probe a full-input model's ability to leverage context to deviate from the predicted partial-input label.}
\label{fig:editing}
\vspace{-1em}
\end{figure}

In this paper, we re-examine such dismissal of the reasoning capabilities of models trained on datasets containing artifacts. While a competitive partial-input baseline is sufficient to cast doubt on the inferential capabilities of full-input models trained on the same data, it is premature to conclude that full-input models are not capable of such reasoning at all. 
As a thought experiment, we imagine training a human to learn a task from a set of examples containing artifacts that allow them to cheat. If we force them to learn to perform the task by removing relevant context, they must rely on the artifacts to accurately perform the task. However, provided all the data, they may still learn to perform the task the ``right way''.

Through two sets of experiments, we investigate whether NLI models are able to condition on the full input despite learning from artifact-ridden datasets. Section \ref{sec:motivation} investigates whether additional context strengthens a full-input model's confidence in the correct label, despite a partial-input model's correct prediction. In Section \ref{sec:editing-experiment}, we introduce an evaluation set to probe whether full-input models are sensitive to changes in context that flip the gold label in examples containing artifacts. We make our evaluation set and code publicly available.\footnote{\scriptsize
 \url{https://github.com/nehasrikn/context-editing}}

Our results indicate that full-input models are still successfully learning to utilize context, overriding strong signal in the partial input. With this finding, we argue that while partial-input baselines are still a useful tool, they do not license the conclusion that models trained on datasets with artifacts do not learn to leverage context from the full input.

\section{Background}
\label{sec:background}

Here we describe the two related tasks of natural language inference (NLI) and \textit{defeasible} NLI; two corresponding datasets (SNLI and $\delta$-NLI, respectively); and annotation artifacts present in these datasets.

\paragraph{Task Definitions.} Natural language inference \cite{giampiccolo2007third, maccartney2009natural, bowman2015large} is the task of determining whether a \textbf{premise} sentence ($P$) entails a \textbf{hypothesis} sentence ($H$). That is, given $P$, would a human conclude $H$ is true (\textit{entailment}), false (\textit{contradiction}), or neither (\textit{neutral}). 

When $H$ is neutral, the task of \textit{defeasible} natural language inference \cite{rudinger2020thinking} asks whether a third \textbf{update} sentence ($U$) makes $H$ \textit{more} likely to be true ($U$ is a \textit{strengthener}) or \textit{less} likely to be true ($U$ is a \textit{weakener}). 
In the following example, $H$ is \textit{neutral} given $P$ (NLI task), and $U$ is a \textit{weakener} given $P$ and $H$ (defeasible NLI task).

\begingroup
\addtolength\leftmargini{-0.2in}
\begin{quote}
\small
    \colorbox{lightteal}{\textcolor{teal}{\textbf{Premise:}}} \hspace{0.1em} A man is sitting in a dim restaurant.
    \colorbox{lightorange}{\textcolor{burntorange}{\textbf{Hypothesis:}}} \hspace{0.1em} He is eating food.\newline
    \colorbox{lightpurple}{\textcolor{darkpurple}{\textbf{Update:}}} \hspace{0.1em} He is browsing a menu. 
\end{quote}
\endgroup

\noindent In this work, we adopt the terms \textbf{context} ($C$) and \textbf{target} ($T$) for clarity when describing partial-input baselines for the two tasks. In NLI, $C=P$ and $T=H$; in defeasible NLI, $C=(P,H)$, and $T=U$. Thus, for either task, the partial-input baseline we use looks only at $T$ and ignores $C$.

\paragraph{SNLI.} SNLI \cite{bowman2015large} is the first large-scale English NLI dataset, containing 570K labeled $P$-$H$ pairs. In SNLI, premises are derived from image captions \cite{young2014image}, and hypotheses for each label (\textit{entailment}, \textit{neutral}, \textit{contradiction}) are elicited from crowdsource workers who are shown a premise.

\paragraph{$\delta$-NLI.} For the task of \textit{defeasible} NLI, \citet{rudinger2020thinking} introduce the $\delta$-NLI dataset, which consists of extensions to three pre-existing English natural language reasoning datasets: SNLI \cite{bowman2015large}, ATOMIC \cite{sap2019atomic}, and SOCIAL-CHEM-101 \cite{forbes2020social}. To extend each dataset, instances (e.g., $P$-$H$ pairs) were presented to crowdworkers who then wrote an update sentence ($U$) to strengthen or weaken the given hypothesis (as described in Task Definitions). The resulting binary classification task is to predict whether an update is a strengthener or weakener, given a $(P, H, U)$ triple. In this work, we focus our evaluation on $\delta$-SNLI, the SNLI portion of the $\delta$-NLI dataset.

\paragraph{Artifacts and Partial-Input Baselines.} \citet{gururangan2018annotation} and \citet{poliak2018hypothesis} observe that the crowdsourcing protocols adopted by \citet{bowman2015large} and others lead to the creation of data with annotation artifacts that enable partial-input baselines (e.g., hypothesis-only baselines) to perform well above a majority-class baseline. For SNLI, \citet{poliak2018hypothesis} report that an InferSent \cite{conneau-etal-2017-supervised} hypothesis-only baseline surpasses a majority baseline by 35 points. A similar effect is observed by \citet{rudinger2020thinking} in the $\delta$-NLI data, with an update-only RoBERTa \cite{liu2019roberta} model achieving 15 points above a majority baseline.

\section{Experiment 1: Context in NLI}
\begin{figure*}
     \centering
     \begin{subfigure}[b]{0.4\textwidth}
         \centering
         \includegraphics[width=\textwidth]{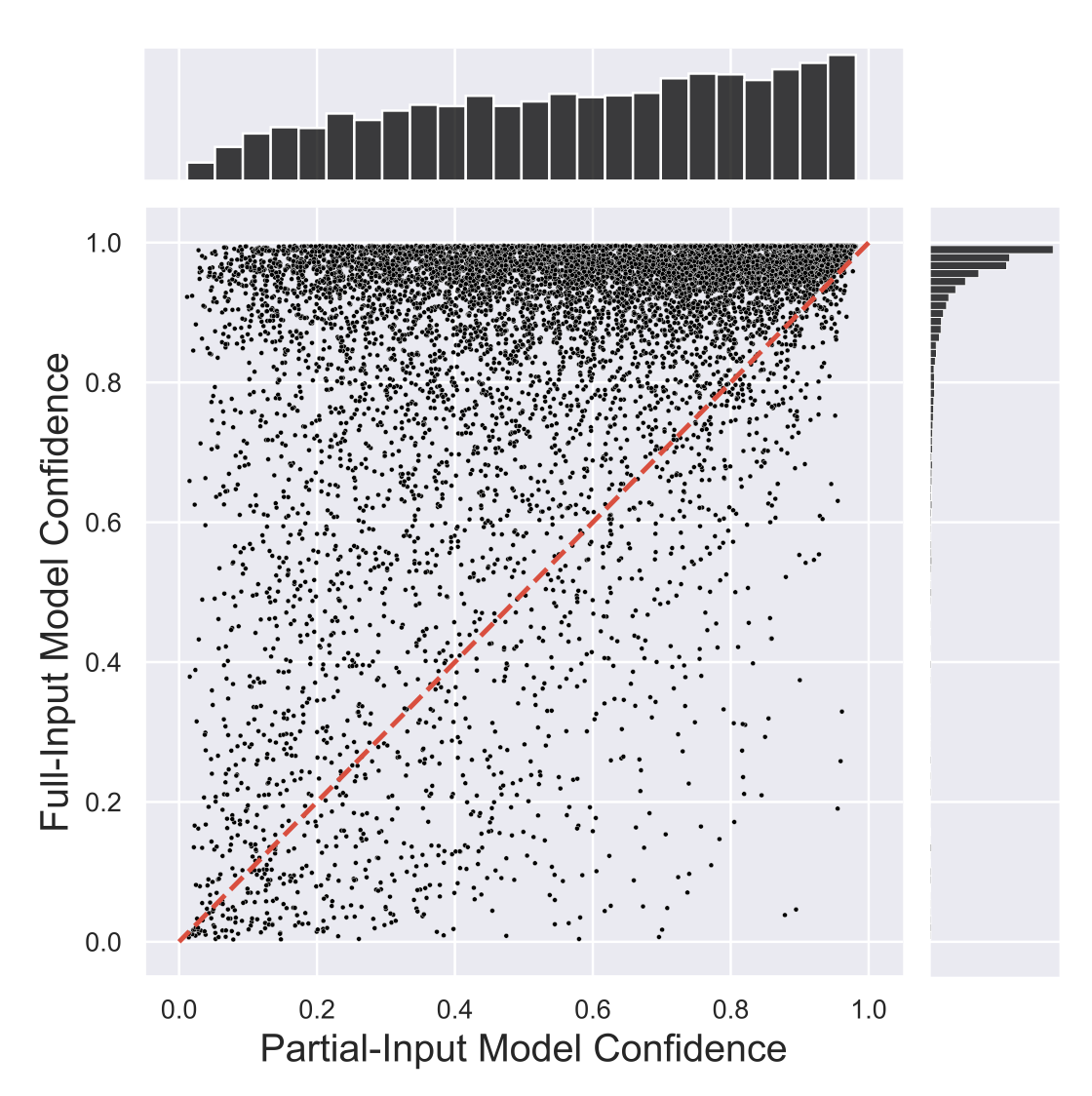}
         \vspace{-1em}
         \caption{Pre-Edit SNLI Confidence Shifts}
         \label{fig:pre-edit-conf-snli}
     \end{subfigure}
     \hspace{1.5em}
     \begin{subfigure}[b]{0.4\textwidth}
         \centering
\includegraphics[width=\textwidth]{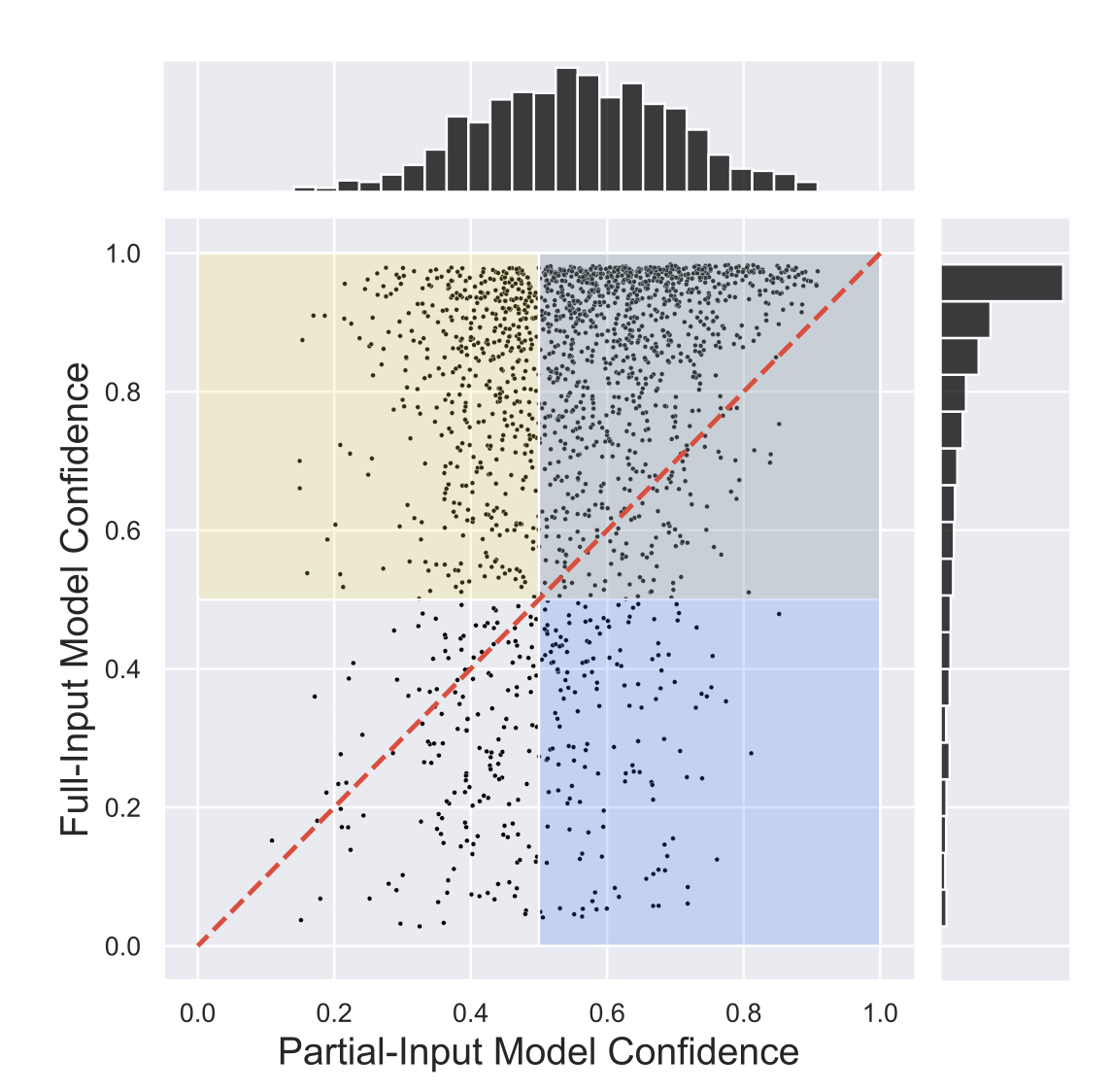}
         \vspace{-1em}
         \caption{Pre-Edit $\delta$-NLI Confidence Shifts}
         \label{fig:pre-edit-conf-dnli}
     \end{subfigure}
     \caption{Visualizing the role of context as confidence shifts. \ref{fig:pre-edit-conf-snli} and \ref{fig:pre-edit-conf-dnli} plot ordered pairs of confidence in the correct label $l$ of partial-input vs. full-input RoBERTa models on examples from the SNLI and $\delta$-SNLI test splits. Blue and yellow regions mark correct predictions from the x and y-axis models respectively. Green regions represent correct predictions from both models. Diagonals indicate no shift in confidence.}
     \label{fig:conf-plots}
     \vspace{-1.5em}
     \hfill
\end{figure*}

An essential component of true reasoning is learning to leverage all parts of an example's input to make a determination of entailment. 
Strong partial-input models demonstrate that full-input models do not necessarily need to utilize context to make correct predictions. 
However, we explore whether they do at all, and how access to such context shifts a full-input model's confidence in the correct label. If, upon supplying $C$, a model strengthens its confidence in its prediction, we may conclude that it utilizes both $C$ and $T$ during inference as intended.

\paragraph{Experimental Setup.} We finetune two sets of RoBERTa \cite{liu2019roberta} models (a partial-input and a full-input model) on the train splits of SNLI and $\delta$-NLI (see Appendix \ref{appendix:dataset-details} for dataset sizes). We utilize \texttt{roberta-base} from the Hugging Face library \cite{wolf2019huggingface}, and finetune each model for two epochs. Appendix \ref{appendix:training-setup} further details our training setup. We then run inference on the test splits from SNLI and $\delta$-SNLI (the SNLI portion of $\delta$-NLI) using each pair of models. Table \ref{tab:test-split-acc} reports accuracy of all models on the corresponding test splits. We calibrate RoBERTa models post-hoc using temperature scaling \cite{guo2017calibration} as suggested by \citet{desai-durrett-2020-calibration}, and examine confidence shifts in the correct label to understand whether full-input models utilize context at all.\footnote{We observed minimal differences between pre and post-calibration results.}

\begin{table}[!t]
\centering
\begin{tabular}{lcc} 
\toprule
 & \multicolumn{1}{l}{SNLI} & \begin{tabular}[c]{@{}c@{}}$\delta$-SNLI\\\end{tabular} \\ 
\hline
\multicolumn{1}{c}{Partial-Input ($C$)} & $0.70$ & $0.65$ \\
Full-Input ($C,T$) & $0.91$ & $0.82$ \\
\bottomrule
\end{tabular}
\caption{Accuracy of both partial-input and full-input models on test splits of SNLI and $\delta$-SNLI.}
\label{tab:test-split-acc}
\end{table}

As shown in Fig. \ref{fig:conf-plots}, we plot an ordered pair of each model's confidence in the \textit{correct label} for examples in the test splits, with the partial-input model's confidence along the x-axis and the full-input model's confidence along the y-axis. Density around the diagonal would indicate no change in confidence.

Evidenced by the density \textit{above the diagonal} in Figures \ref{fig:pre-edit-conf-snli} and \ref{fig:pre-edit-conf-dnli}, full-input models (i.e access to both $C$ and $T$) are more confident in the correct label than partial-input models.\footnote{The distribution difference between SNLI and $\delta$-SNLI may be attributed to the difference in the task difficulty, as well overall lower performance of RoBERTa models on $\delta$-NLI.} While this behavior may seem unsurprising, partial-input baselines illustrate that models may show confidence in the correct label without needing to condition on context at all. Our results hint that full-input models may be successfully learning to leverage additional context instead of overgeneralizing on artifacts in the target. To probe this behavior directly, we introduce an evaluation set crafted by editing contexts in the following section.
\label{sec:motivation}

\section{Experiment 2: Context Editing}
\label{sec:editing-experiment}
We investigate a model's ability to leverage context despite the presence of artifacts by exploring how sensitive full-input models are to changes in \textit{non-target} components of the input. We present an example modification scheme, illustrated in Figure \ref{fig:editing}, in which we edit context sentences from examples where a model correctly predicts the label $l$ from the target $T$ alone. Namely, while holding $T$ constant, we introduce an edited context sentence $C'$ that induces a different label $l'\neq l$ on the new $(C', T)$ pair. Using this scheme, we construct an evaluation set of 600 examples sourced from SNLI and $\delta$-NLI.

\paragraph{Example Subselection.} We select SNLI and $\delta$-SNLI test examples to edit by running the partial-input RoBERTa models from Section \ref{sec:motivation} and full-input bag-of-words (BoW) models, implemented via \texttt{fasttext} with a maximum of 4-grams \cite{joulin2016bag}. See Appendix \ref{appendix:training-setup} for training details on the bag-of-words models. We select examples to edit for which \textit{either} the partial-input model or the BoW model predicted the correct label. This identifies the subset of examples likeliest to contain artifacts in $T$, lexical or otherwise. 

\paragraph{Editing SNLI Examples.} For a given SNLI example $(P, H, l)$ and a new predefined target label $l'$, we edit $P$, creating a modified SNLI example $(P', H, l')$. For each of the six directional pairs of labels (e.g., \textit{entailment} $\rightarrow$ \textit{contradiction}), we randomly sample 50 examples from the subset to edit, resulting in 300 examples evenly distributed across label pairs.

\paragraph{Editing $\delta$-SNLI Examples.} Given a $\delta$-SNLI instance $(P, H, U, l)$, we edit $H$ while holding $P$ and $U$ constant to induce a new label $l'\neq l$, resulting in a modified example $(P, H', U, l')$. We edit 300 examples total, turning 150 \textit{strengthener} examples into \textit{weakener} examples, and vice-versa.

\begin{figure*}[ht!]
\centering
\includegraphics[width=0.86\textwidth]{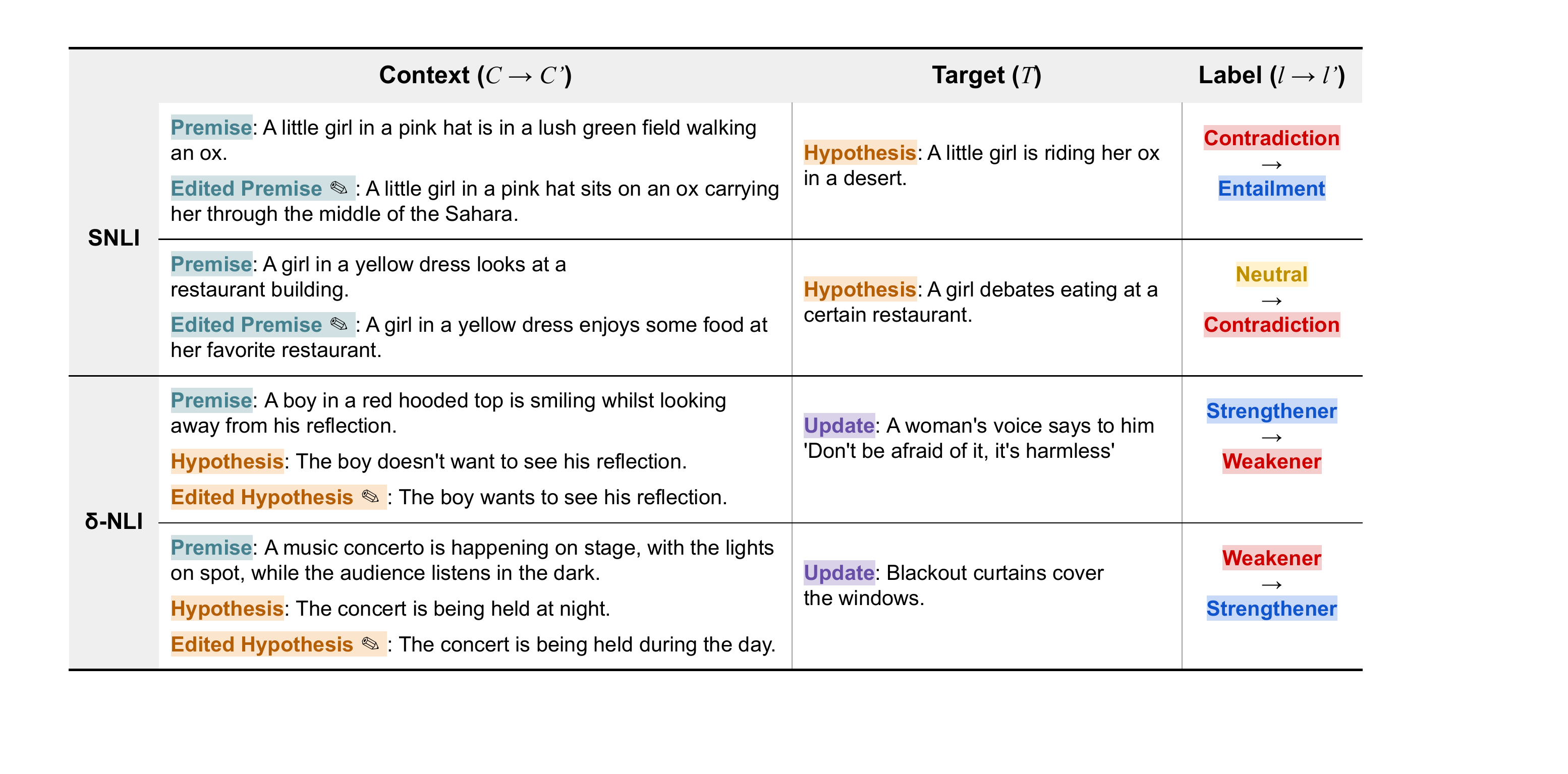}
    \caption{Editing SNLI and $\delta$-SNLI examples. We edit the premise (SNLI) and the hypothesis ($\delta$-NLI), while holding the target constant. The last column contains the original label $l$ and the new induced label $l'$.}
    \label{fig:editing_examples}
\end{figure*}

Our final evaluation set consists of 600 examples containing edited context-target pairs split evenly across SNLI and $\delta$-SNLI. Figure \ref{fig:editing_examples} shows examples of editing contexts from both datasets. All examples were manually edited by one author and independently validated by another. During validation, we hide both the $l$ and $l'$, and ask the annotator to label the text pair. Using Cohen's Kappa \cite{cohen1960coefficient}, we obtain an agreement measure of $\kappa=0.78$ and $\kappa=0.76$ for SNLI and $\delta$-SNLI examples in our test set respectively, indicating substantial agreement \cite{artstein2008inter}.

\begin{table}
\centering
\small
\begin{tabular}{llccc} 
\toprule
 &  & \multicolumn{3}{c}{$l'$} \\
 & \begin{tabular}[c]{@{}l@{}}\\\end{tabular} & entailment & neutral & contradiction \\ 
\cmidrule{2-5}
\multirow{3}{*}{$l$} & entailment & -- & $0.76$ & $0.76$ \\
 & neutral & $0.42$ & -- & $0.78$ \\
 & contradiction & $0.90$ & $0.78$ & -- \\
\bottomrule
\end{tabular}
\caption{A full-input RoBERTa model's accuracy on the edited SNLI portion of our evaluation set. $l$ and $l'$ represent the original and target label respectively, before and after editing.}
\label{tab:snli-perf}
\end{table}

\begin{table}
\centering
\small

\begin{tabular}{llcc} 
\toprule
 &  & \multicolumn{2}{c}{$l'$} \\
 & \begin{tabular}[c]{@{}l@{}}\\\end{tabular} & weakener & strengthener \\ 
\cmidrule{2-4}
\multirow{2}{*}{$l$} & weakener & -- & $0.76$ \\
 & strengthener & $0.75$ & -- \\
\bottomrule
\end{tabular}
\caption{A full-input RoBERTa model's accuracy on the edited $\delta$-SNLI portion of our evaluation set. $l$ and $l'$ again respectively represent the original and induced post-edit label.}
\label{tab:dnli-perf}
\vspace{-5mm}

\end{table}

\begin{figure}[t]
\centering
\includegraphics[scale=0.46]{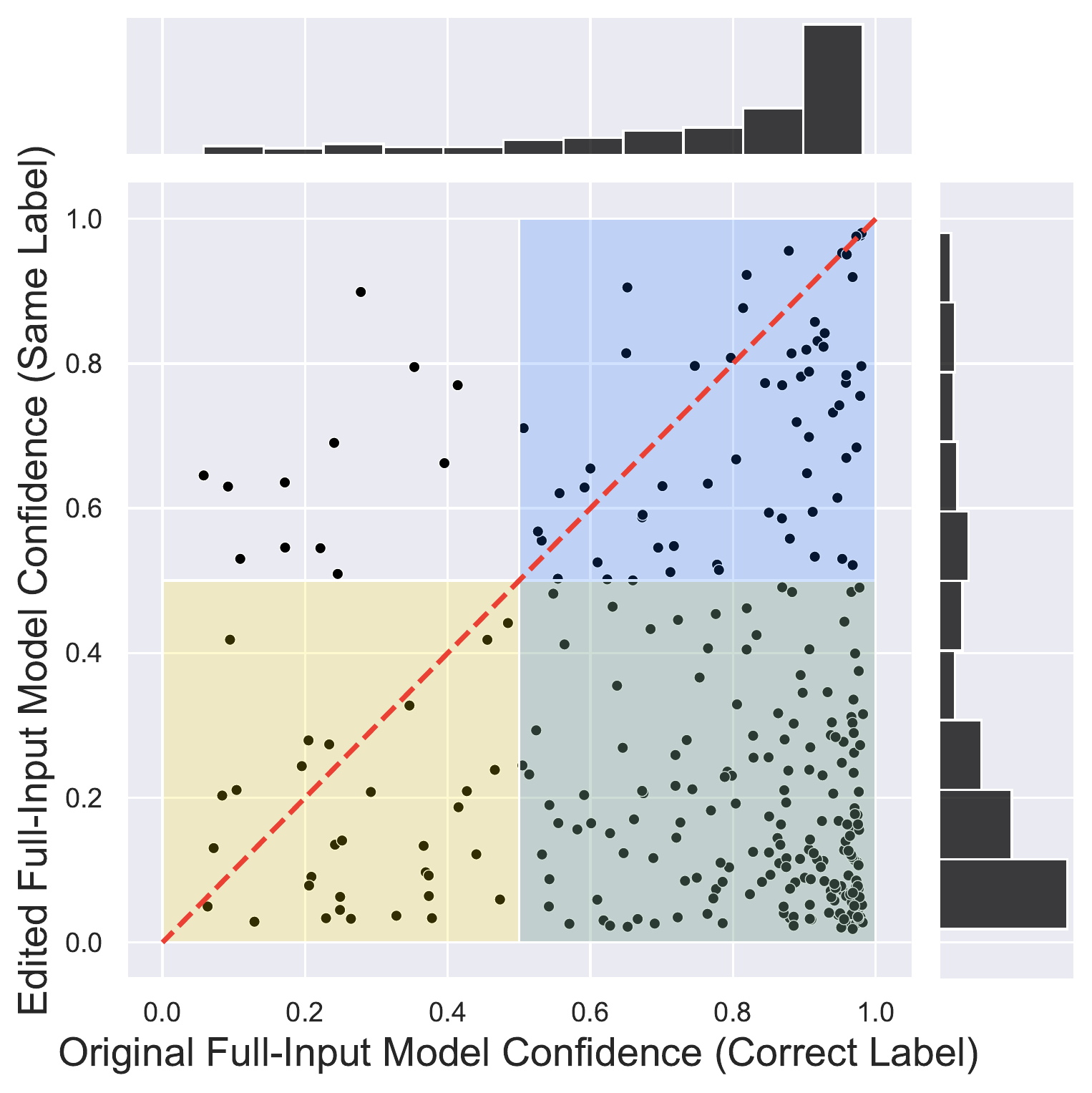}
\caption{Full-input model's confidence in $l$ before vs. after editing $C$ to induce a different label.}
\label{fig:post-edit-conf-dnli}
\vspace{-1em}
\end{figure}

\paragraph{Results.} Using the same full-input RoBERTa models trained in Section \ref{sec:motivation}, we run inference on our edited evaluation set. Tables \ref{tab:snli-perf} and \ref{tab:dnli-perf} show model performance stratified by original label $l$ and target label $l'$. Our results show that full-input models are in fact sensitive to context modifications despite the presence of artifacts in $T$, consistently achieving above 70\% accuracy on edited examples. Thus, we conclude that these models are not overgeneralizing on artifacts in the instance, learning to condition on context for prediction. 

\paragraph{Analyzing Post-Edit Model Confidence.} Similar to the analysis in Section \ref{sec:motivation}, we inspect shifts in confidence upon editing contexts to shed more light on a full-input model's utilization of $C$. For $\delta$-SNLI examples, we plot ordered pairs in Fig. \ref{fig:post-edit-conf-dnli} of a full-input model's confidence in the correct label pre-edit, and its confidence in the \textit{same label} post-edit (i.e the confidence in the now-incorrect label). The majority of mass is under the diagonal, indicating that our model is indeed sensitive to changes in context. The green bottom-left quadrant delineates ideal performance (correct before \textit{and} after editing examples). 
We attribute the small cluster of examples in the blue quadrant (previously highly confident in the correct label and subsequently remained confident, but in the wrong label) to strong, non-lexical artifacts overriding additional signal from the context. Appendix \ref{appendix:simplex} visualizes shifts in SNLI examples using simplex plots to accommodate the ternary label.

\paragraph{Lexical Model Performance.} We do not retrain any of the models on edited examples prior to evaluating with our edited test set, precluding them from picking up on any newly-introduced non-lexical artifacts. However, to validate the absence of trivial lexical features that override the artifacts in $T$ and to ensure sufficient difficulty, we run full-input BoW models on our edited evaluation set. A full-input \texttt{fasttext} model, often used for adversarial filtering \cite{zellers2018swag}, achieves 16\% and 24.3\% accuracy on the SNLI and $\delta$-SNLI portions of our evaluation set respectively.

\section{Related Work}
In addition to work on partial-input baselines for NLI \cite{gururangan2018annotation,poliak2018hypothesis,tsuchiya-2018-performance}, partial-input models have been studied for story completion \cite{cai-etal-2017-pay} and reading comprehension \cite{kaushik-lipton-2018-much}. \citet{feng-etal-2019-misleading} observe that low-scoring partial-input baselines do not preclude other artifacts and heuristics, while \citet{glockner-etal-2018-breaking} and \citet{mccoy-etal-2019-right}  demonstrate lexical and syntactic examples of NLI heuristics. 
Adversarial editing has been explored for non-NLI tasks as well \cite{jia-liang-2017-adversarial,ribeiro-etal-2018-semantically}.
Finally, adversarial filtering has been proposed as a means of removing artifacts from datasets \cite{zellers2018swag,pmlr-v119-bras20a}.

\section{Discussion and Conclusion}
While partial-input models are useful tools for analysis, often leveling fair criticism of datasets, our results show it is hasty to conclude that models trained on such datasets are not capable of reasoning.
Even though high-scoring partial-input baselines show that full-input models \textit{could} ignore context, our experiments show that they can leverage this context quite effectively. 

We argue that artifacts do not \textit{necessarily} spell disaster for a model's reasoning capabilities. In particular, our context-editing experiments identify a set of instances that partial-input models fail (by design), but full-input models largely succeed at, displaying the capability of full-input models to leverage context to overcome SNLI and $\delta$-NLI artifacts in many, but not all, cases.
Of course, we do not deny that artifacts \textit{can} and \textit{do} lead to models with exploitable heuristics, as demonstrated by \citet{glockner-etal-2018-breaking} and \citet{mccoy-etal-2019-right}.

While we do not attempt to define the \textit{sufficient} conditions for a model to perform ``true inference,'' we demonstrate that these models can and do meet the \textit{necessary} condition of leveraging the full input. Thus, we conclude that partial input baselines should be understood as agnostic warning signs: sufficient to conclude that full-input models \textit{might} not be leveraging critical context, but insufficient to prove that they don't.

This raises a number of interesting questions for follow-up work. If adversarial filtering \cite{pmlr-v119-bras20a} can identify instances containing artifacts, is it beneficial to remove these instances from the training set? Or could they be edited to flip the label and mitigate the spurious correlation? The edits we made in this work were done manually, but another interesting question is whether these edits could be made automatically or semi-automatically. Having a more efficient way of producing these examples would enable both rapid evaluation of models trained on datasets with artifacts (as in this work), as well as expansion of training sets to preemptively mitigate artifacts.

\bibliography{acl2021}
\bibliographystyle{acl_natbib}

\clearpage

\appendix
\section{SNLI and $\delta$-NLI Dataset Sizes}
\label{appendix:dataset-details}
To train and tune our neural and lexical models, we utilize the train and validation splits from SNLI and $\delta$-NLI. We include train/validation/test split sizes in Table \ref{tab:data-dist}. Our $\delta$-NLI RoBERTa models were finetuned on examples from all 3 portions of the $\delta$-NLI dataset (ATOMIC, SNLI, and SOCIAL-CHEM-101), however for our evaluation set and analysis, we exclusively use the SNLI portion of the $\delta$-NLI dataset, abbreviated as $\delta$-SNLI. We include the split sizes for $\delta$-SNLI in Table \ref{tab:data-dist} as well.

\begin{table}
\centering
\begin{tabular}{lll} 
\toprule
Dataset & Split & Size \\ 
\hline
\multirow{3}{*}{SNLI} & train & 550,152 \\
 & valid & 10,000 \\
 & test & 10,000 \\ 
\hline
\multirow{3}{*}{$\delta$-NLI} & train & 200,694 \\
 & valid & 14,968 \\
 & test & 15,414 \\ 
\hline
\multirow{3}{*}{$\delta$-SNLI} & train & 88,676 \\
 & valid & 1,785 \\
 & test & 1,837 \\
\bottomrule
\end{tabular}
\caption{Split sizes across SNLI, $\delta$-NLI, and $\delta$-SNLI.}
\label{tab:data-dist}
\end{table}

\section{Model Training Setup}
\label{appendix:training-setup}
\paragraph{Neural Models.} We use the Hugging Face library to train all of our RoBERTa models. We utilize \texttt{roberta-base}, which has 125M trainable parameters. All models were trained on a single NVIDIA 1080-TI GPU. After tuning on the validation set, all models were trained for two epochs with a learning rate of 2e-5 and a batch size of 32. Tables \ref{tab:snli-perf} and \ref{tab:dnli-perf} report best accuracy across five runs. 

\paragraph{Lexical Models.} We use the \texttt{fasttext} library to implement our bag-of-words models. \texttt{fasttext} is an off-the-shelf text classification library. All lexical models were trained for 5 epochs with 4-grams as the maximum length of word ngrams. We use the default learning rate of 0.1.

\section{Visualizing Distribution Shifts in SNLI Edited Examples}
\label{appendix:simplex}
\begin{figure*}
\centering
    \begin{subfigure}[b]{\textwidth}
        \centering
        \includegraphics[width=\textwidth]{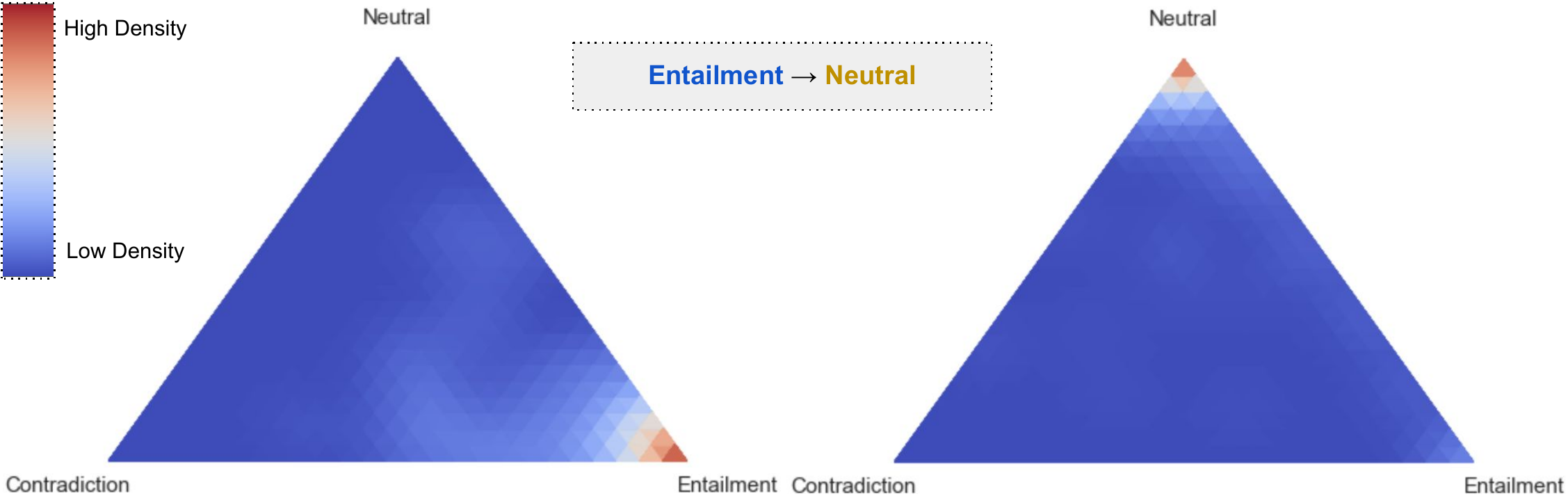}
        \caption{Confidence distribution heatmap for examples with original relation as entailment, edited to induce a neutral relation.}
        \label{fig:snli-e-n}
    \end{subfigure}
    \begin{subfigure}[b]{\textwidth}
        \vspace{3em}
        \centering
        \includegraphics[width=\textwidth]{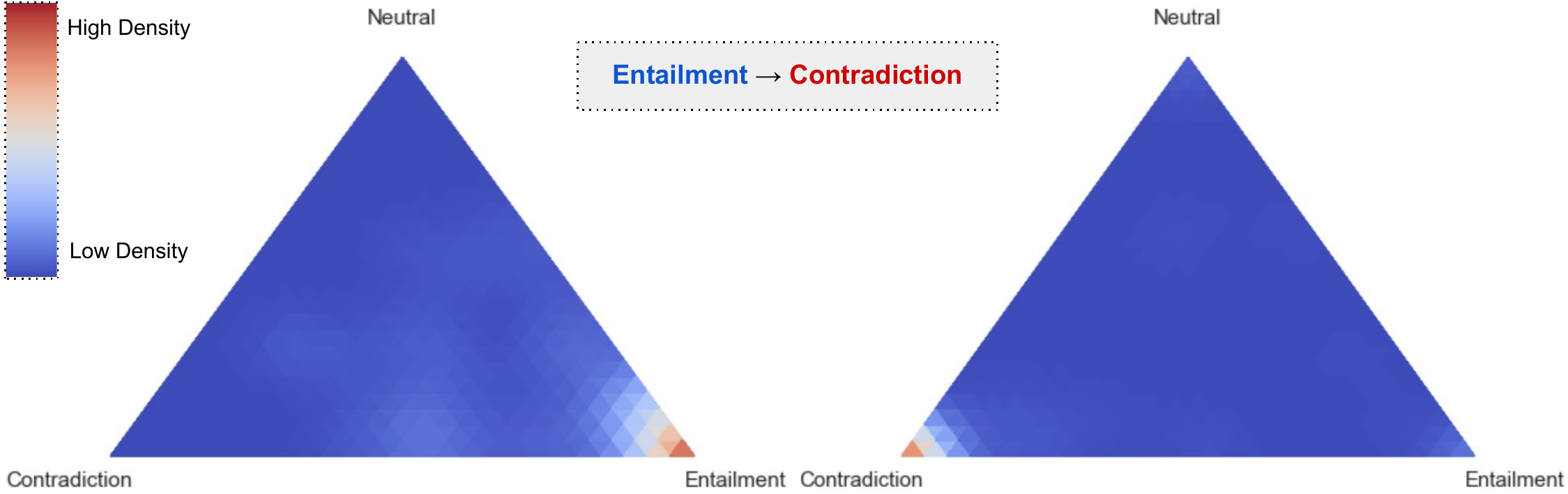}
        \caption{Confidence distribution heatmap for examples with original relation as entailment, edited to induce a contradiction relation.}
        \label{fig:snli-e-c}
    \end{subfigure}
    \caption{Confidence distribution heatmaps for SNLI examples before and after editing examples with original relation as entailment in our evaluation set. \ref{fig:snli-e-n} shows entailment examples edited to induce a neutral relation, and \ref{fig:snli-e-c} shows examples edited to induce a contradiction relation. Each $l$ and $l'$ is shown in the center box. We note that these distributions help visualize the accuracies presented in Table \ref{tab:snli-perf}.}
    \label{fig:entailment-dist-plots}
\end{figure*}

\begin{figure*}
\centering
    \begin{subfigure}[b]{\textwidth}
        \centering
        \includegraphics[width=\textwidth]{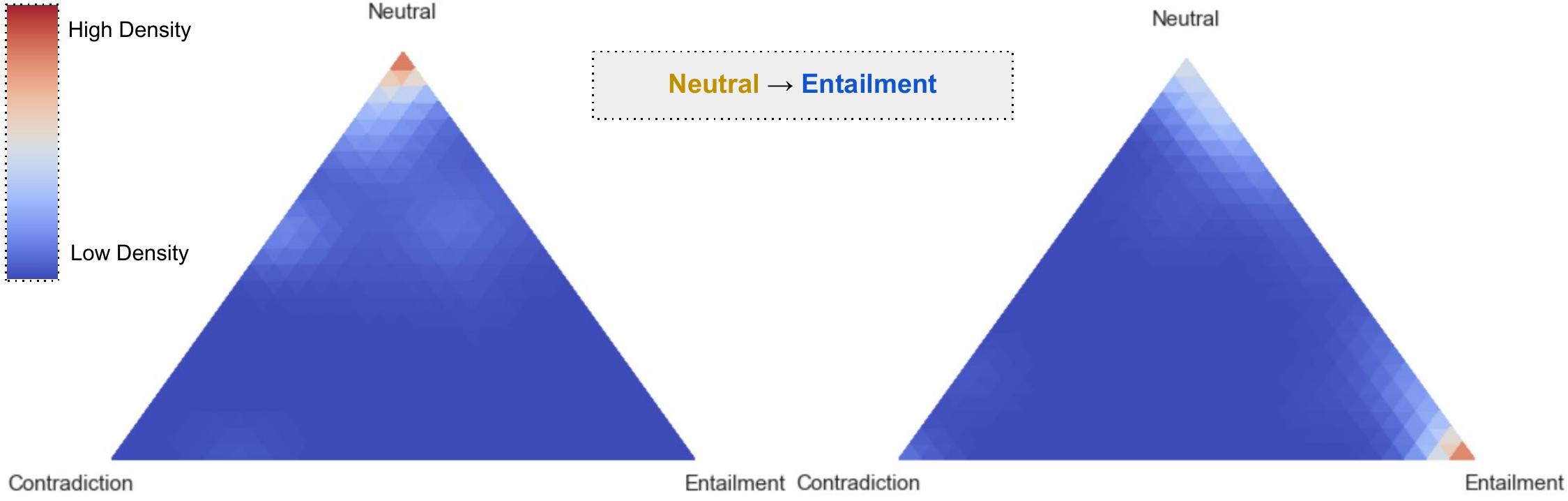}
        \caption{Confidence distribution heatmap for examples with original relation as neutral, edited to induce an entailment relation. We note that this was the hardest edit category for RoBERTa models to flip, and draw attention to a substantial amount of mass still in the neutral corner of the simplex.}
        \label{fig:snli-n-e}
    \end{subfigure}
    \begin{subfigure}[b]{\textwidth}
        \vspace{3em}
        \centering
        \includegraphics[width=\textwidth]{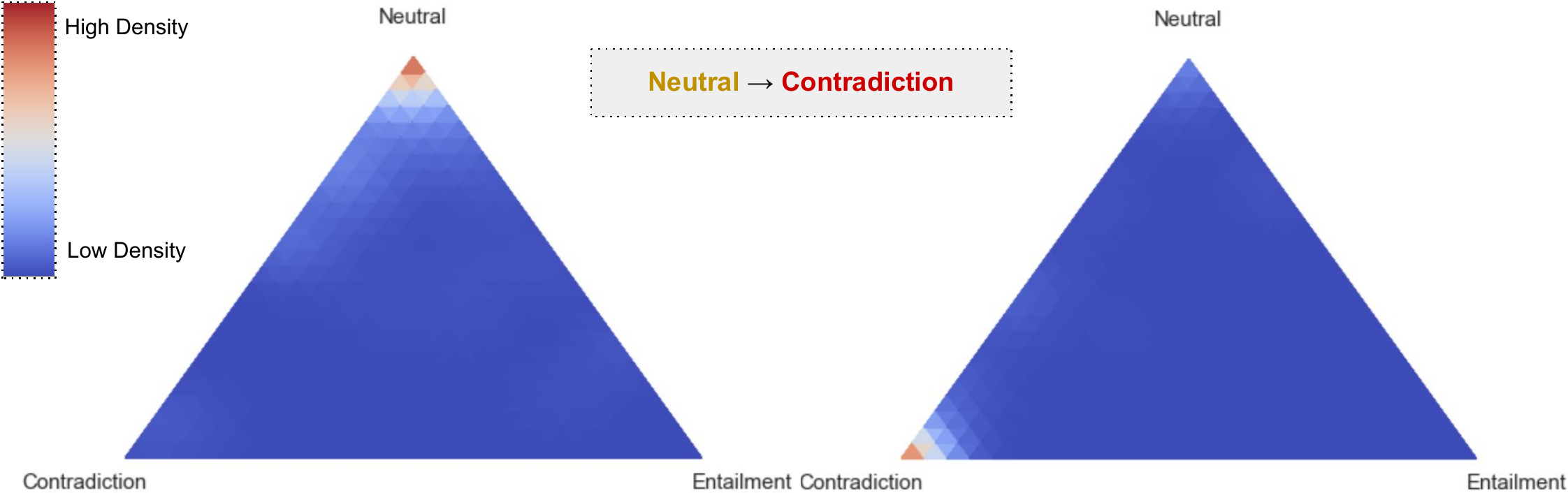}
        \caption{Confidence distribution heatmap for examples with original relation as neutral, edited to induce a contradiction relation.}
        \label{fig:snli-n-c}
    \end{subfigure}
    \caption{Confidence distribution heatmaps for SNLI examples before and after editing examples with original relation as neutral in our evaluation set. \ref{fig:snli-n-e} shows examples edited to induce an entailment relation, and \ref{fig:snli-n-c} shows examples edited to induce a contradiction relation. Each $l$ and $l'$ is shown in the center box. These distributions aid in visualization of the performance metrics reported in Table \ref{tab:snli-perf}.}

    \label{fig:neutral-dist-plots}
\end{figure*}

\begin{figure*}
\centering
    \begin{subfigure}[b]{\textwidth}
        \centering
        \includegraphics[width=\textwidth]{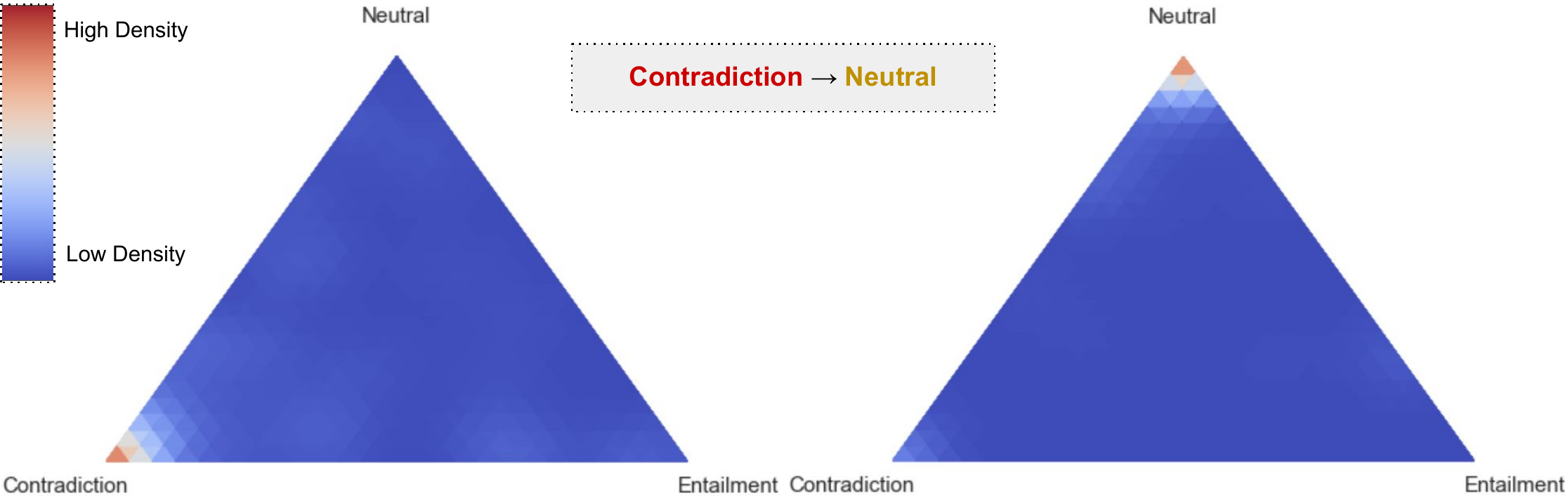}
        \caption{Confidence distribution heatmap for examples with original relation as contradiction, edited to induce a neutral relation.}

        \label{fig:snli-c-n}
    \end{subfigure}
    \vspace{1.5em}
    \begin{subfigure}[b]{\textwidth}
        \vspace{3em}
        \centering
        \includegraphics[width=\textwidth]{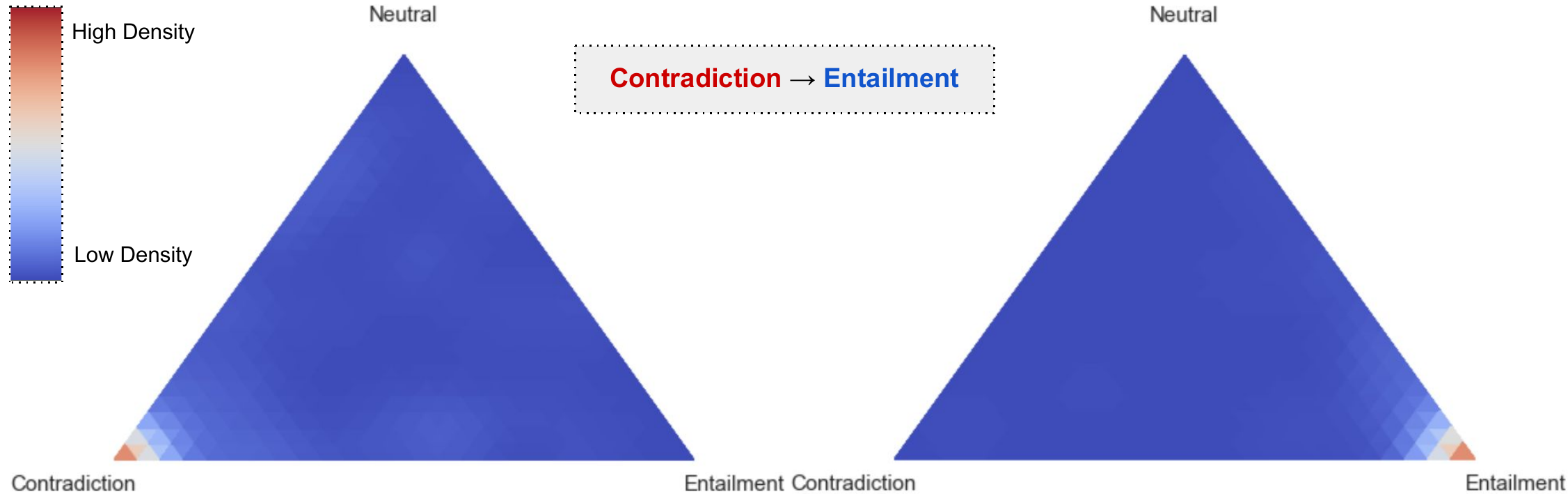}
        \caption{Confidence distribution heatmap for examples with original relation as contradiction, edited to induce an entailment relation. We note that this particular class of edited examples achieved the highest accuracy, reflected in the low density of examples away from the entailment corner of the simplex.}

        \label{fig:snli-c-e}
    \end{subfigure}
        \caption{Confidence distribution heatmaps for SNLI examples before and after editing examples with original relation as contradiction in our evaluation set. \ref{fig:snli-c-n} shows examples edited to induce a neutral relation, and \ref{fig:snli-c-e} shows examples edited to induce an entailment relation. Each $l$ and $l'$ is shown in the center box. These plots help visualize the performance metrics reported in Table \ref{tab:snli-perf}.}
    \label{fig:contradiction-dist-plots}
\end{figure*}

We are able to visualize confidence distribution shifts for $\delta$-NLI before and after editing using a 2D plane with ordered pairs, as in Figure \ref{fig:post-edit-conf-dnli}, due to the label set containing only two update types—weakener and strengthener (so, a probability score $>0.5$ results in the predicted label). Since the SNLI label set consists of \textit{three} labels (\textit{entailment}, \textit{neutral}, \textit{contradiction}), we choose to visualize shifts in the confidence distribution before and after editing contexts via ternary plots. For each directional label pair (\textit{entailment} $\rightarrow$ \{\textit{neutral}, \textit{contradiction}\}, \textit{neutral} $\rightarrow$ \{\textit{entailment}, \textit{contradiction}\}, \textit{contradiction} $\rightarrow$ \{\textit{entailment}, \textit{neutral}\}), we plot a heatmap of probabilities, or confidences, in each of the three classes on the simplex with Gaussian smoothing using \texttt{python-ternary}, a ternary plotting package \cite{pythonternary}. Figures \ref{fig:entailment-dist-plots}, \ref{fig:neutral-dist-plots}, and \ref{fig:contradiction-dist-plots} show these visualizations. We include these plots mainly to help visualize information about the predicted labels of the incorrect examples (Table \ref{tab:snli-perf} only reports the accuracy on finer-grained buckets). We observe that for most classes of examples, the SNLI RoBERTa model utilizes the context, and correctly predicts the new induced gold label. However, the \{\textit{neutral} $\rightarrow$ \textit{entailment}\} class of examples in particular proved difficult for the model, as evidenced by a large chunk of a mass remaining in the neutral corner of the simplex.

\section{Dataset Limitations}
In this work, we choose to explore the role of context with respect to the SNLI and $\delta$-NLI datasets. In particular, the proven presence of strong artifacts in SNLI made it an appealing dataset to explore a model's behavior with respect to the utilization of context. We chose to include $\delta$-NLI in our analysis, since ultimately, we'd like reasoning systems to operate in complex and \textit{dynamic} contexts. The ability to be sensitive to shifting contexts and understand when default inferences should be overridden by additional context (i.e more nuanced inference) is both central to our exploration and central to the task of defeasible reasoning itself. Our evaluation set does not include examples sourced from other NLI datasets such as MultiNLI \cite{williams2017broad}. It also does not contain datasets across domains, such as MedNLI \cite{romanov2018lessons}. However, we note that while the datasets may be different, others have shown artifacts present in such datasets (i.e \cite{herlihy2021mednli, gururangan2018annotation}. Our goal was to utilize datasets containing high amounts of artifacts.

\end{document}